\documentclass[letter, 10pt, conference]{ieeeconf}
\IEEEoverridecommandlockouts

\usepackage{cite}

\usepackage{amsmath,amssymb,amsfonts}
\usepackage{algorithmic}
\usepackage{graphicx}
\usepackage{subcaption}
\usepackage{pdfpages}
\usepackage{textcomp}
\usepackage{xcolor}
\usepackage{bm}
\usepackage{booktabs}
\usepackage{svg}
\svgsetup{inkscapelatex=false}
\usepackage{tikz}
\usepackage{hyperref}
\def\BibTeX{{\rm B\kern-.05em{\sc i\kern-.025em b}\kern-.08em
    T\kern-.1667em\lower.7ex\hbox{E}\kern-.125emX}}
\usetikzlibrary{arrows,decorations.pathmorphing,backgrounds,fit,positioning,calc,shapes}

\usepackage[linesnumbered,ruled,algo2e,resetcount]{algorithm2e}%for the environment
\SetAlFnt{\small}
\SetAlCapFnt{\small}
\SetAlCapNameFnt{\small}

\begin{document}
% IROS LIMIT: 6 pages content

% ------ DECIDE ON A TITLE!
\title{\LARGE \bf
On the programming effort required to generate Behavior Trees and Finite State Machines for robotic applications
\thanks{
\textbf{This work has been submitted to the IEEE for possible publication. Copyright may be transferred without notice, after which this version may no longer be accessible.}
This project is financially supported by the Swedish Foundation for Strategic Research. The authors gratefully acknowledge this support.}
}

%Authors metadata
\author{\authorblockN{
Matteo Iovino$^{a,b}$,
Julian F\"orster$^{c}$,
Pietro Falco$^{a}$,
Jen Jen Chung$^{c,d}$,
Roland Siegwart$^{c}$ and
Christian Smith$^{b}$}\\
\thanks{$^{a}$ABB Corporate Research, Västerås, Sweden}
\thanks{$^{b}$Division of Robotics, Perception and Learning, KTH - Royal Institute of Technology, Stockholm, Sweden}
\thanks{$^{c}$Autonomous Systems Lab, ETH Zürich, Zürich, Switzerland}
\thanks{$^{d}$School of ITEE, The University of Queensland, Australia}
}

\maketitle

\begin{abstract}
In this paper we provide a practical demonstration of how the modularity in a Behavior Tree (BT) decreases the effort in programming a robot task when compared to a Finite State Machine (FSM). 
In recent years the way to represent a task plan to control an autonomous agent has been shifting from the standard FSM towards BTs. Many works in the literature have highlighted and proven the benefits of such design compared to standard approaches, especially in terms of modularity, reactivity and human readability. However, these works have often failed in providing a tangible comparison in the implementation of those policies and the programming effort required to modify them. This is a relevant aspect in many robotic applications, where the design choice is dictated both by the robustness of the policy and by the time required to program it. In this work, we compare backward chained BTs with a fault-tolerant design of FSMs by evaluating the cost to modify them. We validate the analysis with a set of experiments in a simulation environment where a mobile manipulator solves an item fetching task.

\end{abstract}

\begin{keywords}
Behavior Trees, Finite State Machines, Modularity, Mobile Manipulation
\end{keywords}

\section{Introduction}
Nowadays, robots are deployed in unstructured environments, usually shared with humans, where the need for reactivity to handle unpredictable situations and fault-tolerance to increase the robustness of the robot behavior are vital for the successful outcome of the task. Robotics engineers have to take into account these and other features, e.g. programming time, maintainability, readability, when deciding with which policy to control the robot. 
\par
Finite State Machines (FSMs) have been the standard policy representation in robotics due to their intuitive and simple design and the predictability of their execution flow. They have flourished in industrial applications, where the robot repeatedly executes the same task in a static environment with a low failure probability.
As an alternative, Behavior Trees (BTs) are a reactive, readable and modular task switching policy representation, becoming state of the art for robot control, especially in research environments~\cite{iovino_survey_2022}.
\par
We believe that examples of how these two representations compare on a concrete implementation are still lacking. Therefore, the goal of this paper is to focus the comparison between BTs and FSMs on the modularity of the two representations, also carrying out experiments in a simulation environment. In this work, we emphasize modularity since it is a key feature that allows users to drastically reduce the programming effort required for a robotics application, especially when it is possible to reuse existing code or libraries. First, we measure modularity in terms of Cyclomatic Complexity (CC),  Edit Distance (ED), and Computational Complexity of the operations required to edit the policy. Then, we show the practical implications of choosing one representation rather than the other with a series of simulated robotic tasks.

\begin{figure}[t]
    \centering
    \includegraphics[width=.7\linewidth]{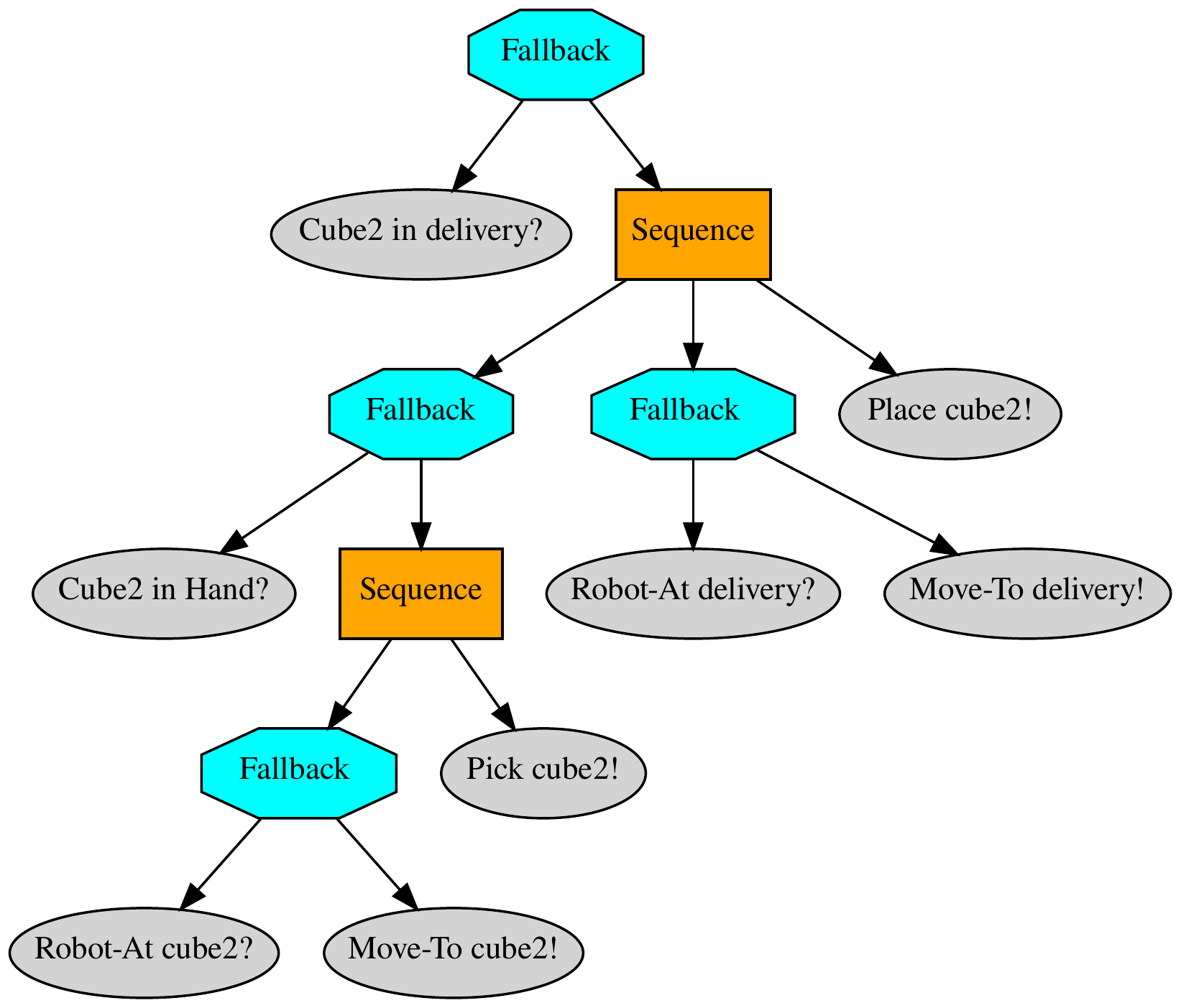}
    \caption{Backchained BT solving a mobile manipulation task.}
    \label{fig:base_BT}
\end{figure}

To make the comparison as fair as possible, we make a few assumptions on the design:
\begin{enumerate}
    \item The two representations have access to the same robot skill set, so that the difference remains at as high a level as possible;
    \item We design BTs with the backward-chained (backchained) principle because it allows deriving stability guarantees and convergence proofs~\cite{ogren_convergence_2020}, even though it is not the most optimal choice in terms of the size of the tree (as is found by e.g. Genetic Programming algorithms~\cite{iovino_learning_2021, styrud_combining_2022});
    \item We propose to build FSMs in a fault-tolerant way, so that status checks on the environment redirect the robot to the point in the execution flow that had previously failed. This mimics the reactiveness of BTs. We validate this choice in a direct comparison with a standard sequential FSM.
\end{enumerate}

For implementation, we use state-of-the-art open-access programming tools\----BTs: \texttt{py\_trees}\cite{py_trees}; FSMs: \texttt{SMACH}~\cite{bohren_smach_2010}\cite{smach}\----which are widely used in robotic applications~\cite{ghzouli_behavior_2022}.

\section{Background and Related Work}
This section briefly defines BTs and FSMs and presents the theoretical foundations and early results on the comparison between the two representations.

\begin{figure}[tbp]
    \centering
    \includegraphics[height=5cm]{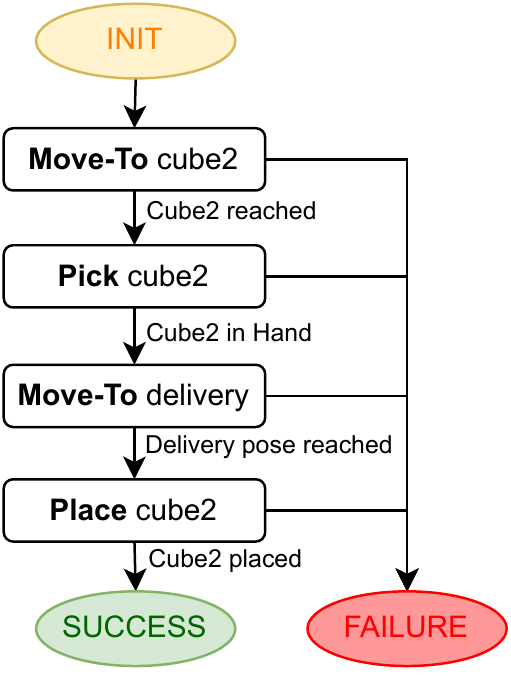}
    \caption{Sequential FSM solving the same task of Fig.~\ref{fig:base_BT}.}
    \label{fig:seq_sm}
\end{figure}

\subsection{Behavior Trees}
BTs are a task switching policy representation, originating in the gaming industry and later transferred to robotics~\cite{colledanchise_behavior_2018}.
\par
A BT is a directed tree, where a tick signal originates from the root and propagates down the tree from left to right. Nodes execute only when ticked and return one of the status signals \textit{Success}, \textit{Failure} and \textit{Running}. Internal nodes are called \emph{control nodes}, (polygons in Fig.~\ref{fig:base_BT}), with the most common types being \emph{Sequence}: runs children in a sequence, returning once all succeed or one fails, and \emph{Fallback} (or \emph{Selector}): runs children in a sequence, returning when one succeeds or all fail. Leaves are called \emph{execution nodes} or \emph{behaviors} (ovals in Fig.~\ref{fig:base_BT}), of type Action(!) or Condition(?). The former encode robot skills, while the latter encode status checks and sensory readings, thus immediately returning \textit{Success} or \textit{Failure}. 
\par
BTs have explicit support for task hierarchy, action sequencing and reactivity~\cite{iovino_survey_2022} and are modular by design: every node receives the tick as input and outputs the return statuses, so subtrees can be moved without compromising the logical functioning. Moreover, modularity allows every building block to be independently tested and reused. Reactivity is realized by the \emph{Running} return state, which allows high priority actions to preempt executing ones.
\par
A backchained BT is built from the goal condition, by expanding it with the actions that achieve it and then by recursively expanding those actions' unmet pre-conditions. As an example, in the BT of Fig.~\ref{fig:base_BT}, placing the cube in the delivery station allows the robot to complete the task, but in order to reach that stage, the robot first has to pick the cube and then move to the delivery station. Recursively, to pick the cube, the robot has to first move to where the cube is located. The backchained design can also be automatically generated by a planner~\cite{styrud_combining_2022, colledanchise_towards_2019} or learnt~\cite{gustavsson_combining_2022, iovino_interactive_2022}, making it appealing to operators with limited programming knowledge.

\begin{figure}[tbp]
    \centering
    \includegraphics[height=5cm]{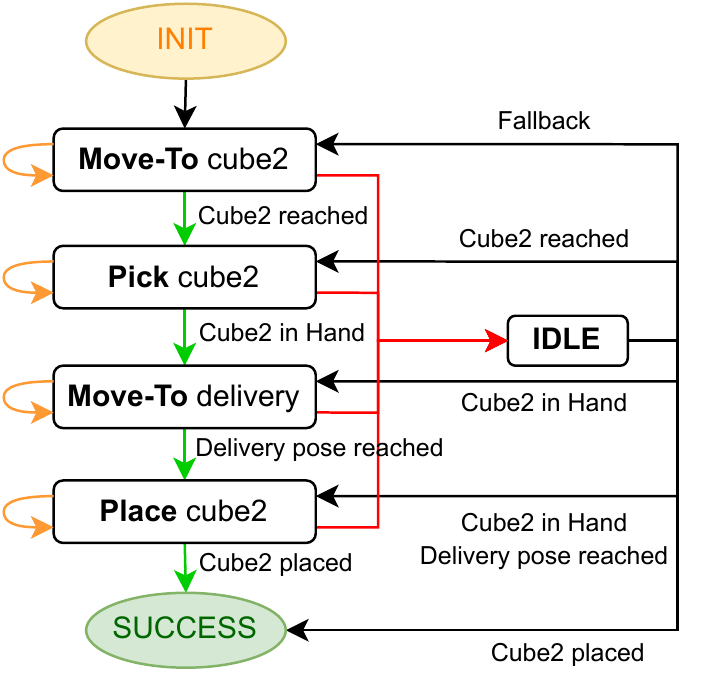}
    \caption{Fault-tolerant design for FSMs solving the same task of Fig.~\ref{fig:base_BT}. States' outcomes \textit{Success}, \textit{Running} and \textit{Failure} have green, yellow and red transitions, respectively.}
    \label{fig:fault_sm}
\end{figure}

\subsection{Finite State Machines}

Finite State Machines derive from state automata and feature a set of states and transitions between them (ovals and arrows in Fig.~\ref{fig:seq_sm}, respectively). Every state encodes a controller for robot behavior which produces effects in the environment upon execution. The effects trigger an event that transfers the execution from one state to the next. Since FSMs also include Sequential Function Charts (SFCs), a graphical programming language for Programmable Logic Controllers (PLCs), they are widely used in industry~\cite{lepuschitz_using_2019} and their success is mainly due to their intuitive design and implementation simplicity.
\par
FSMs have the unfortunate shortcoming that designers have to make a trade-off between reactivity and modularity.
As highlighted in~\cite{colledanchise_behavior_2018}, FSM execution can be compared to the GoTo statement of early programming languages, where the execution flow jumps from one part of the program to another and continues from there. In programming, and by extension also in robotics, GoTo statements are considered to be harmful and are advised against~\cite{dijkstra_letters_1968}. On the other hand, the execution of a BT can be compared to a function call, where the execution flow also jumps to another part of the code but after its completion, it returns to where the function was initially called.
\par
In order to be reactive, an FSM needs to have many transitions that need to be taken care of upon addition or removal of states, making them less modular and less scalable. The modularity problem is partially mitigated by logically grouping states to form hierarchies. There is a particular design for Hierarchical State Machines (HFSMs) aimed to mimic the functionality of a BT and to grant FSMs some modularity~\cite{colledanchise_how_2017}. Since this design sacrifices the readability, we leave the analysis of the modularity of HFSMs to future work.
\par
Another shortcoming of sequential FSMs (e.g. Fig.~\ref{fig:seq_sm}) is that the robot and the environment have to be reset upon failure, and the task to be re-executed again from the start. To make the FSM fault-tolerant, we propose to add an \textit{IDLE} state that is fully connected to the others (as in Fig.~\ref{fig:fault_sm}). In case of failure of any of the action states, the execution flow passes to the IDLE state, where the status of the environment and the task progress is checked, allowing the robot to resume the execution in the correct state. We endow every state with an explicit \emph{Running} transition that cycles back to the state itself, so that the action execution is asynchronous and the environment is monitored periodically. An action's post-conditions (also referred to as effects) are the transitions to the next state (or from IDLE to another state), while its pre-conditions are reflected in the sequential order of the states in the FSM. Note that an alternative fault-tolerant and reactive design would feature a task switcher node to which all other states are connected and which is responsible for checking the state of the environment and deciding the next sequence of actions. In this design, however, we lose the sequentiality of the execution flow and, consequently, we also lose the readability as it is harder for a human operator to anticipate what the robot will do next.

\subsection{Related Work}

BTs have been compared to FSMs in previous work, but the comparison was either purely theoretical~\cite{colledanchise_behavior_2018, colledanchise_how_2017, biggar_expressiveness_2021}, or speculative~\cite{klockner_behavior_2013}. In~\cite{colledanchise_behavior_2018}, authors list advantages and disadvantages of both designs and in~\cite{colledanchise_how_2017} they prove theoretically how in fact BTs modularize FSMs, providing an HFSM design that behaves like a BT. In~\cite{biggar_expressiveness_2021}, authors formally compare BTs with other related architectures (Decision Trees, Teleo-reactive Programs and in particular FSMs) in terms of reactivity, readability and expressiveness.
\par
In~\cite{biggar_modularity_2021}, the same authors formalize modularity for reactive control architectures, pointing out that BTs feature structural interfaces, with which \textit{every component} interacts with the others: a subtree is a BT, an action behavior is a degenerate case of a BT. Naturally, FSMs lack structural interfaces and thus cannot be considered modular in their analysis unless a structure is enforced (e.g. in HFSMs). They measure modularity with the Cyclomatic Complexity (CC) that we will describe later in the analysis. In addition we propose to measure modularity in terms of the effort required to modify a structure by adding or removing elements in it. We propose to quantify it using the Computational Complexity of such operations and the Edit Distance (ED) between a baseline structure and its modified versions. 
\par
Klöckner~\cite{klockner_behavior_2013} proposes using a BT as a control policy for UAV missions, speculating the advantages of such design with respect to FSMs.
Previous work~\cite{olsson_behavior_2016} also compares the two policy representations from a practical perspective in the domain of autonomous driving. The comparison is made in terms of CC and Maintainability Index (MI). The MI of a piece of software takes into account the CC, the number of lines of code, the percentage of comments in the code and the Halstead Volume (a function of distinct and total numbers of operands and operators). Here, the CC was ill-defined as compared to~\cite{biggar_modularity_2021} and we argue that the MI is more dependent on the library used to implement the two policies, so we disregard it.
\par
Another work that provides insight on the behavior of an agent controlled by both policies is~\cite{colledanchise_learning_2018} in the domain of computer games, i.e. in a deterministic scenario. Here, the authors evaluate the policies on solving a level in the Mario AI benchmark, comparing them in terms of number of nodes and reward function $\rho(x)$. This comparison provides insight on the scalability of the two policies, since both of them are generated by a learning algorithm to solve the benchmark. The goal of this work was to propose a new method to generate BTs, so the comparison against FSM is not fair in the sense that the generation method is not the same for the two policies.
In both~\cite{olsson_behavior_2016} and~\cite{colledanchise_learning_2018} the complexity scales linearly for BTs and quadratically or worse for FSMs.
From an application point of view, \cite{colledanchise_implementation_2021} provides guidelines to program BTs for robotic tasks, in terms of parameter passing, asynchronous calls to robot actions, and reactivity, which are also applied here in the experimental section. 
\par
We believe that what is lacking in the state of the art is a practical grasp on the difference between BTs and FSMs. Therefore, we propose to support the analytical comparison with a simulated robotic application, where we provide insights on the programming effort required to design both structures and comment on which one may be preferable over the other.

\section{Analytic Comparison on Modularity}

\begin{figure*}[tbp]
     \centering
\begin{subfigure}[b]{0.45\textwidth}
    \centering
    \includegraphics[width=.4\linewidth]{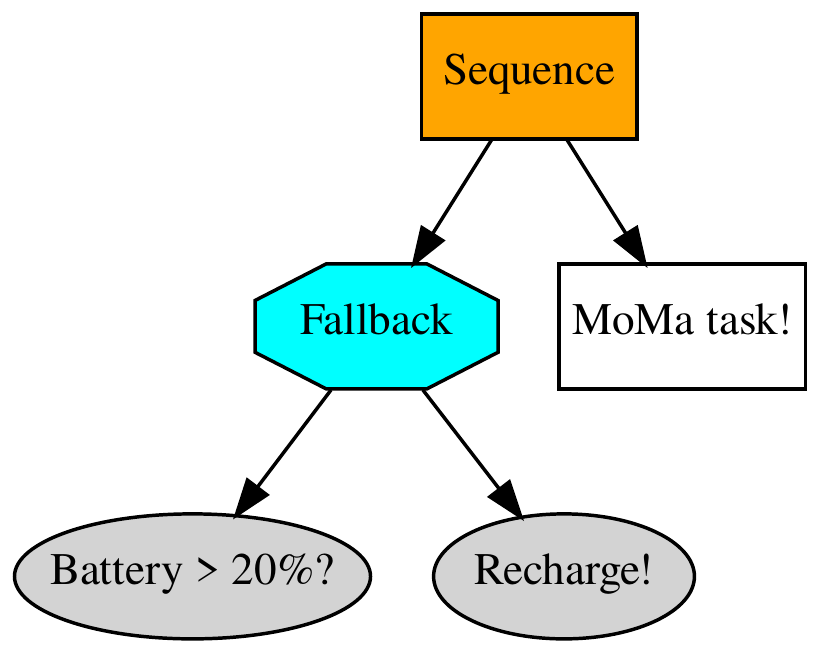}
    \caption{Adding a subtree that recharges the battery if low.}
    \label{fig:recharge_BT}
\end{subfigure}
\hfill
\begin{subfigure}[b]{0.45\textwidth}
    \centering
    \includegraphics[width=.4\linewidth]{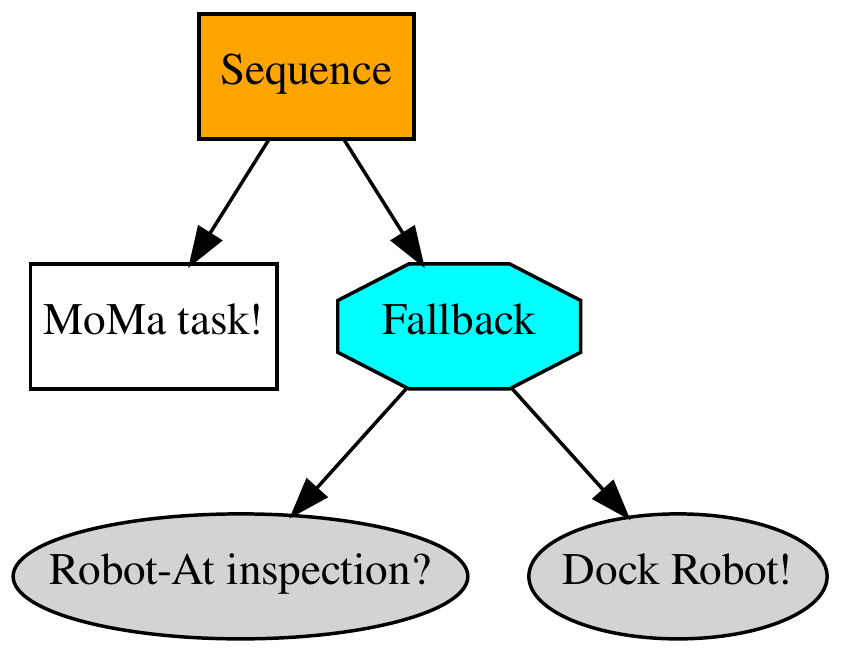}
    \caption{Adding a state that docks the robot once the task is done.}
    \label{fig:dock_BT}
\end{subfigure}
\hfill
\begin{subfigure}[b]{0.55\textwidth}
    \centering
    \includegraphics[height=5.5cm]{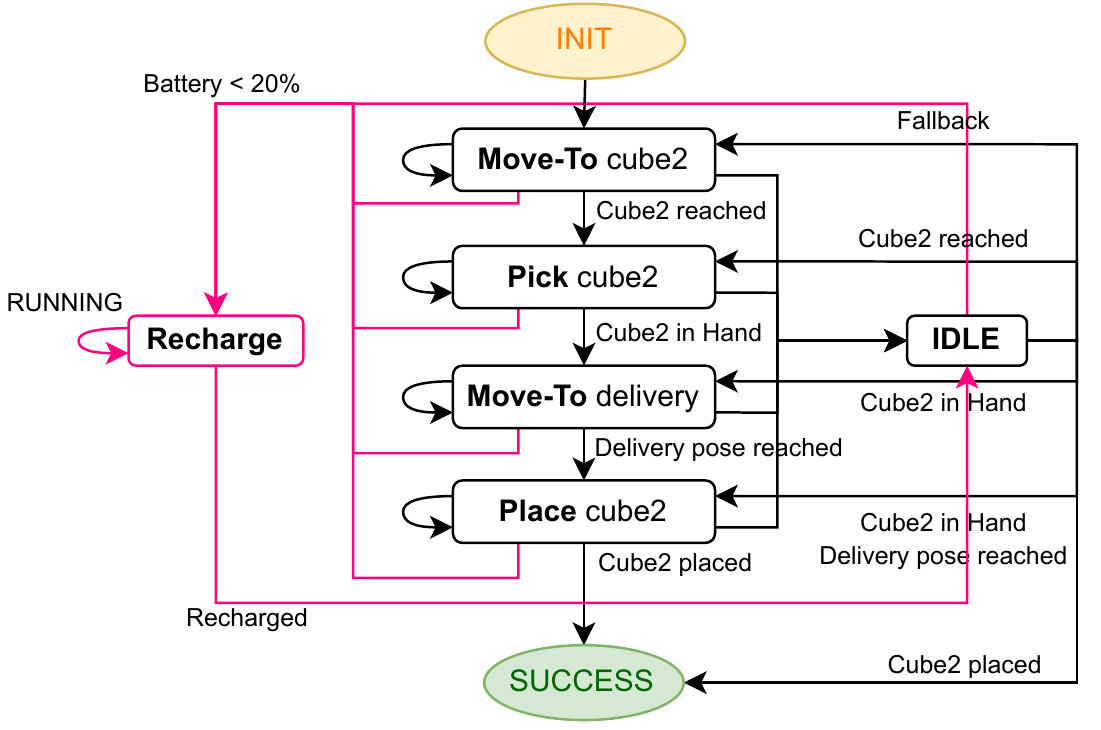}
    \caption{Adding a state that recharges the battery if low.}
    \label{fig:recharge_SM}
\end{subfigure}
\hfill
\begin{subfigure}[b]{0.35\textwidth}
    \centering
    \includegraphics[height=5.5cm]{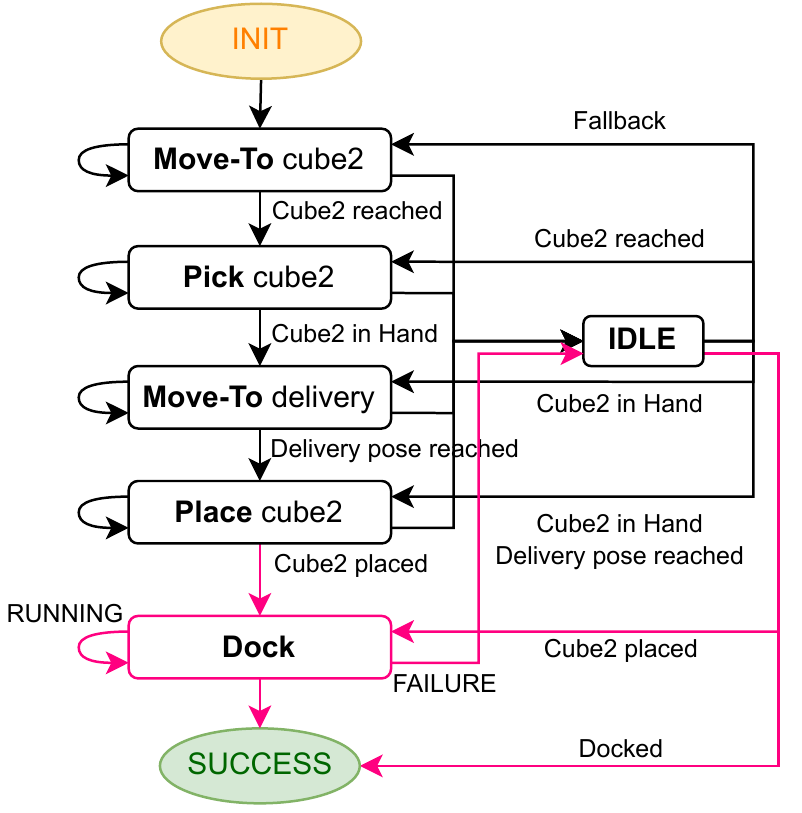}
    \caption{Adding a state that docks the robot.}
    \label{fig:dock_SM}
\end{subfigure}
\caption{How the addition of a new behavior is handled in the two policy representations. To save space, we collapsed the BT in Fig.~\ref{fig:base_BT} in the white box labelled \textit{`MoMa task!'}. In the FSM, additions are colored in magenta.}
        \label{fig:add}\vspace{-0.65cm}
\end{figure*}

We use the Computational Complexity, Edit Distance (ED), and Cyclomatic Complexity (CC) to compare the modularity of the two representations. 
\par
The possibility of adding and removing nodes/states allows users to reuse pieces of software and to modify bad design choices. It can happen, for example, that during the testing of the policy we realize that some behaviors were missing or that some were malfunctioning, and thus need to be removed.

\subsubsection{\textbf{Computational Complexity}}

\paragraph{Add or remove a node in a BT} Adding a node in a BT with $n$ nodes, requires inserting it in the list of children of the parent control node at the desired position. In a BT, each child node is in no way connected to other children\footnote{While not true in the case of using Blackboard variables, this design choice breaks modularity and it is discouraged~\cite{colledanchise_behavior_2018}.}. Performing the insertion operation requires accessing the desired parent node and then inserting it, with complexity $\mathcal{O}(1)$. Removing a node has the same complexity.
Because of modularity, handling a subtree has the same complexity. In this case, the root of the subtree to add/remove is considered as the input node. 

\paragraph{Add or remove a node in a FSM} For the proposed design of the FSM, we identify two possible ways a state can be added:
\begin{enumerate}
    \item[i)] \textbf{sequential state}: add a new state between two existing states (namely the \textit{preceding} and \textit{following} states). If we want to add the new state as a new step in the execution sequence, we need to remove the transition from \textit{preceding} to \textit{following}, create a transition from \textit{preceding} to the new state and from this to \textit{following}. Finally we need to handle the interaction with the IDLE state. If the new state is a terminal one, then we would make a transition to the outcome instead. A FSM modified this way is reported in Fig.~\ref{fig:dock_SM}.
\end{enumerate}
\begin{enumerate}
    \item[ii)] \textbf{connected state}: add a new state for the task that is connected to all the others (Algorithm~\ref{alg:sm_add3}). Here, we add a transition from every other state to the new one and finally to handle the interaction with the IDLE state. A FSM modified this way is reported in Fig.~\ref{fig:recharge_SM}.
\end{enumerate}

\begin{algorithm2e}[t]
\caption{FSM: Addition of a connected state}\label{alg:sm_add3}
% make the package mimic python code
\SetStartEndCondition{ }{}{}%
\SetKwProg{Fn}{def}{\string:}{}
\SetKw{KwTo}{in}\SetKwFor{For}{for}{\string:}{}%
\SetKwIF{If}{ElseIf}{Else}{if}{:}{elif}{else:}{}%
\SetKwFor{While}{while}{:}{fintq}%
\AlgoDontDisplayBlockMarkers\SetAlgoNoEnd\SetAlgoNoLine%
% user defined keywords and functions
\SetKwData{Cond}{condition}
\SetKwData{ICond}{idle\_condition}
\SetKwData{IDLE}{IDLE}
\SetKwData{State}{state}
\SetKwData{FSM}{sm}
\SetKwFunction{Outcome}{register\_outcome}
\SetKwFunction{AddTr}{add\_transition}
\SetKwFunction{AddSt}{add\_state}
\SetKwInOut{Input}{input}
% pseudocode
\Input{sm, new\_state, condition, idle\_condition}
\BlankLine
\For{state \KwTo \FSM}{
    $state$.\Outcome(new\_state)

    \If{node == \IDLE}{
        $node$.\AddTr(\ICond)
    }
    \Else{
        $node$.\AddTr(\Cond)
    }
}
$new\_transitions \gets "RUNNING":\ new\_state$

$new\_transitions \gets "FAILURE":\ $\IDLE

\FSM.\AddSt($new\_state$, $new\_transitions$)
\end{algorithm2e}

% Since the FSM keeps a register of all the transitions in the form of a dictionary, adding any transition is an operation with complexity $\mathcal{O}(1)$, but removing a specific one is $\mathcal{O}(n)$, because we need to take care of $n$ transitions.
In a FSM with $n$ states, adding or removing a new state requires checking the consistences between state outcomes and transitions, which has complexity $\mathcal{O}(n)$ because we might need to take care of up to $n$ transitions.
\par
In particular, to remove a state, it is necessary to delete all transitions to and from the state. Then, the target state has to be removed from the outcomes of any other state, if present. This operation requires looping through all transitions and all states, for a total complexity of $\mathcal{O}(2n)$.
\par
A fundamental difference with BTs\----a direct consequence of BT modularity\----is that while editing a FSM it is necessary to have access to all states and transitions, making the operation of adding/removing one element in the structure dependent on all the others.

\subsubsection{\textbf{Edit Distance (ED)}}

ED is a way to measure diversity between structures. With the formulation proposed in~\cite{burke_diversity_2004}, two BTs to compare are padded with empty nodes into the same shape.
Since this formulation of the ED is defined for trees alone, we cannot directly apply it to FSMs. Using the fact that trees are a special type of graph, we propose instead to use the Graph Edit Distance (GED). GED is defined as the minimum number of edit operations (add/remove/substitute nodes and edges) to execute on a graph $g_1$ to make it isomorphic to another graph $g_2$~\cite{abu-aisheh_exact_2015}. Using the standard tuple definition of graphs, $G = (V, E)$ with $V,E$ the sets of vertices and edges respectively, the GED between $g_1 = (V_1, E_1)$ and  $g_2 = (V_2, E_2)$, is defined as:
\begin{equation}
    GED(g_1, g_2) = \min_{e_1,\dots,e_k\in \gamma(g_1, g_2)} \sum_{i=1}^k c(e_i),
\end{equation}
where $c$ is the cost of the edit operation $e_i$ and $\gamma(g_1, g_2)$ denotes the set of edit paths to transform one graph into the other. For this analysis, we use this measure as implemented in the \texttt{NetworkX} Python library~\cite{hagberg_exploring_2008}.

%\paragraph{Edit distance in BTs}

%Considering BTs as a graph, we have the following:
%\begin{itemize}
%    \item the BT in Fig.~\ref{fig:base_BT} has 14 nodes and 13 edges;
%    \item the BTs~\ref{fig:recharge_BT} and~\ref{fig:dock_BT} have 18 nodes and 17 edges.
%\end{itemize}
%We obtain an edit distance of 8. The best scenario possible, where just a single node is added, has $ED=2$.

%\paragraph{Edit distance in FSMs}

%We use the GED to compare FSMs of Fig.~\ref{fig:add} with the baseline of Fig.~\ref{fig:fault_sm}. Such FSMs are directed graphs with the following elements:
%\begin{itemize}
%    \item the FSM in Fig.~\ref{fig:fault_sm} has 6 nodes and 18 edges;
%    \item the FSM in Fig.~\ref{fig:recharge_SM} has 7 nodes and 25 edges;
%     \item the FSM in Fig.~\ref{fig:dock_SM} has 7 nodes and 22 edges;
%\end{itemize}
%where the outcome \textit{Success} is also considered as a node. We obtain EDs of $8$ and $5$ respectively.

%According to the GED metric the two designs are comparable, although the computational complexity to modify a FSM is notably higher than for the BT case. Note that in the BT example, a subtree is added because of the backchained design that requires linking an action's pre- and post-conditions. Moreover, in both cases the BT root is also created. If it were already there, we would have an ED of $6$.

\subsubsection{\textbf{Cyclomatic Complexity (CC)}}
This is a software metric that measures the complexity of a program in the form of a control-flow graph and is defined as,
 \begin{equation}
     CC = a+s-n+1,
 \end{equation}
where $a, s, n$ represent the number of arcs, sinks (terminal nodes) and nodes in a decision structure, respectively. This measure is applied in~\cite{biggar_modularity_2021} to BTs that have been transformed to graphs with single entry and exit nodes. This implies that \textbf{BTs have optimal modularity} as their Cyclomatic Complexity is 1.
%In the fault-tolerant FSM designs of Figs~\ref{fig:fault_sm},\ref{fig:recharge_SM} and \ref{fig:dock_SM} we have $CC=\{14,20,17\}$, respectively.

%The results of the case study are summarized in Table~\ref{tab:properties}.

%\begin{table}
%\centering
%\caption{Overview on the modularity of the policies according to the metrics. In a BT $n$ is the %number of nodes and in a FSM $n$ is the number of states and transitions.}
%\begin{tabular}{c c c c}
%\toprule
%Policy &%\multicolumn{1}{p{2cm}}{\centering Computational \\ 
%Complexity & \{min, max\} ED & CC \\
%\midrule
%BTs & $\mathcal{O}(1)$ & 2, 8 & 1\\
%FSMs & $\mathcal{O}(2n)$ & 5, 8 & 14, 20, 17\\
%\bottomrule
%\end{tabular}
%\label{tab:properties}
%\end{table}

\begin{figure}[t]
    \centering
    \includegraphics[width=.7\linewidth]{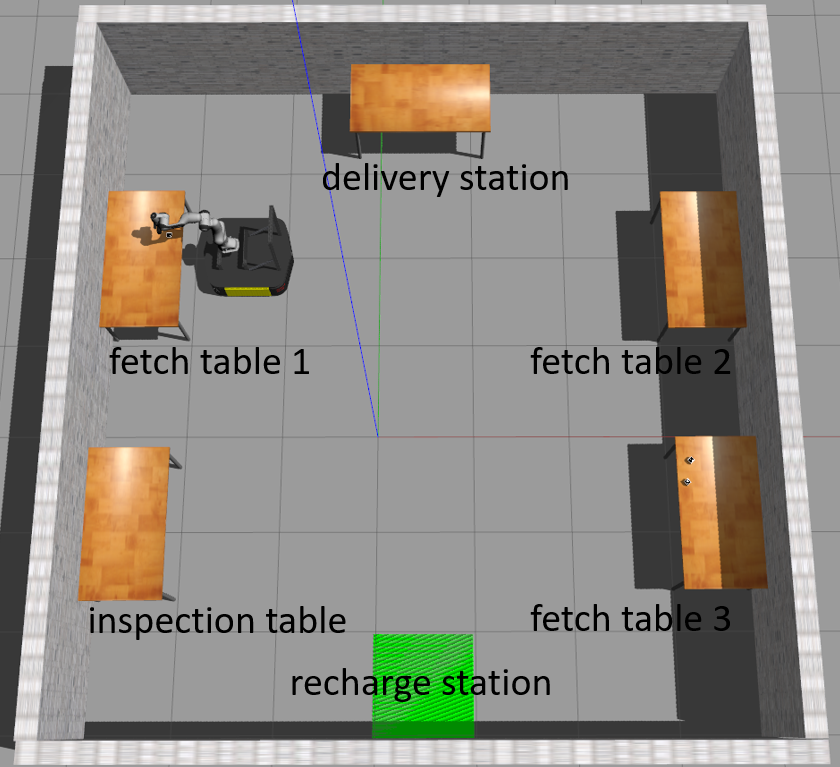}
    \caption{Gazebo Simulation environment with the mobile manipulator.}
    \label{fig:sim}
\end{figure}

\section{Experiments}

We compare BTs and FSMs in the concrete case of controlling a robot to perform a simulated mobile manipulation (MoMa) task and we evaluate the metrics described above on some specific edits. The MoMa task shown in Figs~\ref{fig:base_BT} and~\ref{fig:fault_sm} is used as the baseline.
We target tasks that are typical and representative of the MoMa domain, not to bias the analysis in favor of one policy or the other. To this end, we consider a set of variations of the \textbf{Cleaning Up} task as defined in the RoboCup@Home benchmark~\cite{wisspeintner_robocuphome_2009}. In this task a robot needs to collect a set of five unknown objects dispersed throughout the scene. To do so, the robot has to find the potential object, grasp it, and finally place it in a predefined area in the scenario. The map is known in advance. To perform this task the robot needs functional abilities such as navigation, object recognition, and object manipulation, as well as system properties such as adaptivity (the object locations are not known in advance and the environment is unpredictable), robustness, and general applicability.
\par
We consider the simpler subtask of fetching one single object of known position, as it facilitates the comparison and reduces the execution time. 
The evaluation is divided into pairs of experiments, where the task becomes more complicated as new behaviors are added to the policy. The task is solved once with a BT and once with a FSM. We aim at studying the behavior of the robot as well as the effort of programming such a policy, in terms of the number of operations required. The code is available online\footnote{\url{https://github.com/ethz-asl/bt_fsm_comparison}}.
To complexify the task, we add functionalities according to the following scenarios, Fig.~\ref{fig:add}:
\begin{itemize}
    \item The robot must recharge its batteries if they run low at any point during task execution.
    \item The robot must dock once the task is completed.
\end{itemize}
\par
In the simulation environment for the task (Fig.~\ref{fig:sim}), the robot docks at the \emph{inspection table}. The mobile manipulator is a Franka Emika Panda arm mounted on the omnidirectional Clearpath Ridgeback base. The arm's end-effector is equipped with an Intel RealSense RGBD camera.

\subsection{Skill Description}
Since the scope of the paper is to compare the high-level behavior of the control policies, we make some assumptions to simplify the execution of the low-level skills. In the case of the navigation, the robot has a limited set of target poses available (one for each station/table of the environment). The navigation is implemented with the ROS Navigation Stack.
%, which can behave unpredictably when the robot is close to obstacles, e.g. the tables in the experiment domain. For this reason, we decided to us the Navigation stack to position the robot $0.5m$ away from the obstacle, followed by covering the remaining distance using open-loop velocity commands.
We use AprilTags to identify the location of the items.
%For perception, we use AprilTags to identify the location of the items, which is then stored and available to the robot at anytime.
Manipulation is implemented with ROS MoveIt!. For grasping, we rely on a shape-based grasp synthesis method called the Volumetric Grasping Network (VGN)~\cite{breyer2020volumetric}.

In the ROS framework, skills are implemented as action servers. A client interface allows the policy (BT or FSM) to interact with the skills, i.e. to initialize them, send goals, monitor the execution status and cancel goals. 
In the following, we describe the skills and the conditions available to the robot.
\paragraph{\textbf{Move-To!}} this action moves the robot to a target pose given explicitly or reconstructed from the object marker.  
\paragraph{\textbf{Robot-At?}} this condition checks that the robot reached the desired pose with the desired tolerance in x, y and yaw.
\paragraph{\textbf{Pick!}} this action moves the robot arm to a target pose and grasps the object. Once the grasp is successful, the arm is tucked in a compact configuration for navigation.
\paragraph{\textbf{In Hand?}} this condition checks if the robot is holding an object.
\paragraph{\textbf{Place!}} this action moves the robot arm to a target pose and drops the object. Then the robot moves the arm to a configuration that allows it to monitor the scene and evaluate the end pose of the object with the wrist camera.
\paragraph{\textbf{Object-At?}} this condition checks that the object is in the desired pose with the desired tolerance in x, y, and z.
\paragraph{\textbf{Recharge!}} this action makes the robot go to the recharge station and then it instantaneously fills up the battery level.%, represented as a counter that decreases constantly with time. 
\paragraph{\textbf{Battery Lv?}} this condition determines when the robot shall recharge its batteries. %, with a user-selected threshold and a comparison sign (lower/greater) as parameters.
%\paragraph{\textbf{Search!}} this action makes the robot follow 5 viewpoints (one per table in Figure~\ref{fig:sim}), where the robot stops and inspects the table with the camera to look for markers \todo{check if used}. 
%\paragraph{\textbf{Found?}} this condition determines which markers to search for and by consequence, when to stop the search behavior \todo{check if used}.
\paragraph{\textbf{Dock!}} this action makes the robot dock at the \emph{inspection table}. The corresponding condition is `Robot-At'?

\subsection{Experiment 1}
In this experiment the robot has to fetch the cube placed at \emph{`fetch table 1'} in Fig.~\ref{fig:sim} and place it on the \emph{`delivery station'}.
The robot starts in the center of the room and the position of the cube is known a priori. This is the only task that is also attempted with the sequential FSM of Fig.~\ref{fig:seq_sm}. The BT and fault-tolerant FSM solving the task are those of Fig.~\ref{fig:base_BT} and  Fig.~\ref{fig:fault_sm}, respectively. The sequential FSM successfully solves the task only in the case all actions execute correctly and within the tolerances. Failures in the execution require the task to be reset. As per the BT and the fault-tolerant FSM, failed actions are re-attempted. In addition, since the BT is recursively ticked from the root, it keeps being executed also upon success. This implies that if the the task is completed but then a human operator brings the cube to another table, the robot can react to it and go fetch the cube again (provided it knows where the cube is). To calculate the ED when adding behaviors, for this base task, the BT in Fig.~\ref{fig:base_BT} is a graph with 14 nodes and 13 edges, while the FSM in Fig.~\ref{fig:fault_sm} has 6 nodes and 18 edges. This FSM has $CC=14$.

\subsection{Experiment 2}
This experiment adds a recharging behavior to the existing manipulation task. This requires the operator to edit the policy as in Fig.~\ref{fig:recharge_BT} and  Fig.~\ref{fig:recharge_SM}. In the case of the BT we would need to prepend the recharge subtree in a \emph{Sequence} root node to the subtree solving the task, so the recharging subtree has higher priority. There are 8 elementary operations to perform: create the 4 nodes\----the root node and the nodes in the left subtree\----in Fig.~\ref{fig:recharge_BT}, add the two leaves to the Fallback node and finally add the recharge subtree and the already existing BT for the MoMa task to the new root.
\par
Note that it is assumed that the recharge behavior loads the batteries instantaneously. This can be motivated by a situation in which for example, an operator switches the batteries of the robot when it stops at the recharge station. In the case where the batteries are gradually loaded when the robot stays in the recharge station, the recharge subtree could be given a different design to preempt charging if a more important task arises.
\par
The BTs in Fig.~\ref{fig:recharge_BT} has 18 nodes and 17 edges, which infers an ED of 8 with respect to the baseline.
\par
In the case of the fault-tolerant FSM, there are also 8 operations to perform: create the recharge state and the \textit{Running} transition, add a transition from the recharge state to the IDLE state and from every state to the recharge state. There are two considerations to make. First, the number of new transitions to create depends on the number of already existing nodes, while in the case of the BT, the number of operations required does not depend on the size of the tree. Then, the internal logic of the states must also be modified to implement the case triggering a transition to the new state. This is a clear benefit of a BT, where the switching logic is explicitly implemented in the representation. 
\par
The FSM in Fig.~\ref{fig:recharge_SM} has 7 nodes and 25 edges, which infers an ED of 8 with respect to the baseline, and the new FSM now has $CC=20$.
\par
From a behavioral perspective, the execution of the task remains the same and the robot successfully goes to recharge the batteries if the level is below $20\%$.

\subsection{Experiment 3}
In this case, we also add a docking behavior as in Figs~\ref{fig:dock_BT} and~\ref{fig:dock_SM}. Since the \textit{Sequence} root node is already there, to edit the BT it is necessary (1) to create the \textit{Fallback} node, the condition and the action, (2) to add the two leaves to the control node, and (3) to add the subtree to the root. Again, the rest of the tree is in no way influenced by the addition.
\par
In the FSM, besides creating the new state, the \textit{Running} transition and the transitions to and from the IDLE state, we need to remove the transition from the \emph{`Place'} state to the \textit{Success} outcome and create a transition from the \emph{`Dock'} state to the outcome.
\par
In this experiment as well, there is no substantial difference in the executed robot behavior between the two policies. The BT has 21 nodes and 20 edges, which gives $ED=6$ to the previous case of \textit{Exp.2}, while the FSM has 8 states and 30 transitions which gives $ED=6$ and $CC=24$. The figures for this examples are reported in the code repository.

\subsection{Scalability}
For the experiments, we have considered only one item to fetch. While a task that simple allows us to detail the implementation and programming efforts, it doesn't fully capture the real advantages that modularity gives to the BT when compared to FSMs. We have commented that insertion operations on FSMs depend on the number of states and transitions already in the FSM. In fact, when we scale the task to a case where the robot needs to search for 5 cubes, to fetch all of them and finally to dock, we would have a BT with 77 nodes and 76 edges and a FSM with 24 nodes and 90 transitions, given the set of skills that we presented. At this point, adding a recharge behavior to the existing policy representations would be more cumbersome in the case of a FSM as we need to add the transition to the recharge state from all other states. We would have $ED=6$ in the case of a BT (like in \textit{Exp.3}, the root is already there) but $ED=26$ in the case of a FSM, as the final BT would feature 80 nodes and 79 edges, while the final FSM 25 nodes and 115 transitions.

%\begin{itemize}
 %   \item run the base experiment in case of sequential FSM, fault tolerant FSM, BT;
 %   \item show that sequential FSM is not fault tolerant and needs to be restarted upon failure;
 %   \item show that BTs and fault tolerant FSMs are indeed fault tolerant.
  %  \item show that if the task becomes more complex (adding a recharge behavior), then BTs are easier to handle than FSMs;
   % \item other comments: the BT is more explicit so the logic is already in the representation, while in a FSM the logic is inside the state and hence the robot behavior is more difficult to understand;
%\end{itemize}

\section{Conclusions and Future Work} \label{sec:conclusion}

In this paper we analyzed modularity in BTs and FSMs in terms of Computational Complexity, Edit Distance, and Cyclomatic Complexity. In order to ensure that the comparison is between representations with equivalent functionalities, we used a particular design of the FSM that gives it fault-tolerance. This is also motivated by the fact that BTs are a structured policy representation and there are implementation guidelines, while the design of a FSM is less constrained. Moreover, we built the BTs with the backchained design for the discussed reasons. The results of the theoretical analysis highlight how BTs are indeed more modular than FSMs, as already identified by related work~\cite{colledanchise_how_2017, biggar_modularity_2021}. We contributed with a detailed analysis on the implementation of the two policies and deployment on a set of MoMa tasks in a simulation environment. From the experimental section, it is undeniable that FSMs are more intuitive to implement as the functioning and state-transitioning logic is more straightforward. However, it is more complicated to maintain a FSM and the number of operations needed to edit the policy depends on the number of states in the structure. Moreover, it is often necessary to change the logical behavior of some states if the task requires it. Modularity allows us to easily maintain, reuse, and extend BTs, as edit operations depend only on the parent node of the subtree to modify. Another remark is that the switching policy is more explicit in a BT, while in a FSM it is integrated within the state. As a future work, we will continue this analysis with an extended set of experiments where we aim to find the size of the policy (in number of states/action nodes) where the trade-off between ease of implementation and maintainability tips the balance in favor of BTs.
Moreover, we plan to extend the comparison to other key elements for policy representation in robotics, e.g. reactivity and readability.

\section*{Acknowledgments}
Authors would like to thank Michel Breyer, Julian Keller and Yinyin Liu from the Autonomous Systems Lab, ETH Zürich, for their support in implementing the robot skills.

\bibliographystyle{IEEEtran}
\bibliography{references.bib}

\end{document}